%% file: main.tex
\documentclass[journal]{IEEEtran}
\usepackage[T1]{fontenc}
\usepackage{tgtermes}
\usepackage{amsmath,amssymb}
\usepackage{graphicx}
\usepackage{booktabs}
\usepackage{array}
\usepackage{xcolor}
\usepackage{microtype}
\usepackage{cite}
\usepackage[hidelinks]{hyperref}
\input{numbers}
\input{numbers_lto}

\newcommand{\nesy}{\textsc{Halal-NeSyNet}}
\newcommand{\ahr}{\mathcal{L}_{\mathrm{AHR}}}
\newcommand{\sv}{\mathbf{s}}

\hypersetup{pdftitle={Halal-NeSyNet: Clause-Scoped Neuro-Symbolic Cross-Attention for Automated Food Ingredient Auditing and Risk-Aware Halal Logistics Triage},
  pdfauthor={Mochamad Rizal Fauzan, Nai'la Nursyifa, Irgi Surya},
  pdfkeywords={Halal supply chain; Halal logistics; food additives; E-numbers; neuro-symbolic AI; cross-attention; counterfactual evaluation; warehouse triage}}

\begin{document}

\title{Halal-NeSyNet: Clause-Scoped Neuro-Symbolic\\
Cross-Attention for Automated Food Ingredient\\
Auditing and Risk-Aware Halal Logistics Triage}

\author{%
\begin{tabular}[t]{@{}c@{}}
Mochamad Rizal Fauzan\\[3pt]
{\small National Taipei University of Technology}\\
{\small Taipei, Taiwan}\\
{\small t114998415@ntut.edu.tw}
\end{tabular}
\hspace{1.0cm}
\begin{tabular}[t]{@{}c@{}}
Nai'la Nursyifa\\[3pt]
{\small Logistics Engineering}\\
{\small Universitas Pendidikan Indonesia}\\
{\small Bandung, Indonesia}\\
{\small nailanrsy@upi.edu}
\end{tabular}
\hspace{1.0cm}
\begin{tabular}[t]{@{}c@{}}
Irgi Surya\\[3pt]
{\small National Taipei University of Technology}\\
{\small Taipei, Taiwan}\\
{\small t115999404@ntut.edu.tw}
\end{tabular}}

\markboth{Fauzan \MakeLowercase{\textit{et al.}}: Halal-NeSyNet}{Fauzan \MakeLowercase{\textit{et al.}}: Halal-NeSyNet}

\maketitle

\begin{abstract}
Verifying that a packaged food is Halal requires reading ingredient declarations and E-number additives before goods enter certified warehouses.
Dual-origin additives such as E471, E422, E441 and E920 are \emph{Mushbooh} (doubtful) unless a botanical, microbial or certified-Halal qualifier appears in the same ingredient clause, so a plant-based qualifier on one clause must not clear an unqualified additive in another.
We call a failure to respect this scope \emph{cross-clause modifier confounding} and build a benchmark of \NItemsFmt\ items (\NReal\ Open Food Facts records and \NSynFmt\ template-generated clause-swap items, labelled by an explicit \NEcode-code rule base rather than by a certification authority) with a \NCf-item counterfactual suite.
We propose \nesy{}, which couples a DistilBERT encoder with a 16-dimensional rule-derived symbolic vector through cross-attention and gated fusion, trained with an asymmetric false-Halal risk loss.
Over five seeds, text-only TF-IDF baselines reach \TxtAccMin--\TxtAccMax\,\% accuracy with \TxtFhlrMin--\TxtFhlrMax\,\% false-Halal leakage and \CfTxtTriadMin--\CfTxtTriadMax\,\% counterfactual triad consistency, whereas \nesy{} reaches \MsNsAccPm\,\% and \CfNsTriadPm\,\%.
Gradient boosting given the same symbolic vector reaches \MsXgbOAccPm\,\% ($\PdNsXgbOAcc$ points from \nesy{}, 95\,\% CI [$\PdNsXgbOAccLo$, $\PdNsXgbOAccHi$]), so most of the gain comes from clause-scoped symbolic features and not from the network, and the risk loss did not help.
Holding out whole template families, text-only models fall to \LtoTxtAccMin--\LtoTxtAccMax\,\%, \nesy{} to \LtoNsAccPm\,\% and logistic regression with the symbolic vector to \LtoLrOAccPm\,\%, with far more spread for \nesy{}.
A three-lane receiving simulation auto-clears \TrNsGreen\,\% of cargo at \TrNsCont\,\% green-lane contamination.
Because labels and features come from one rule engine and \PctSyn\,\% of the items are templated, the results measure fidelity to a rule policy, not religious certification.
\end{abstract}


\input{sections/intro}
\input{sections/related}
\input{sections/data}
\input{sections/method}
\input{sections/results}

\input{sections/discussion}
\input{sections/conclusion}

\section*{Data and Code Availability}
The scripts for data extraction, preprocessing, training, multi-seed evaluation, the re-run of the ontology-augmented baselines and the generation of every table, figure and number in this paper accompany the submission; the result files, the benchmark and the counterfactual suite can be obtained from the authors on request, and a public repository will accompany a later version.
The real declarations were retrieved from the Open Food Facts API in October 2026 \cite{openfoodfacts}.

\input{references}
\end{document}

%% file: numbers.tex
\newcommand{\AbNoAhrAccPm}{97.54$\pm$0.40}

\newcommand{\AbNoAhrFhlrPm}{4.59$\pm$1.09}

\newcommand{\AbNoCaAccPm}{96.25$\pm$0.87}

\newcommand{\AbNoCaFhlrPm}{6.41$\pm$1.24}

\newcommand{\CfDbAccPm}{99.22$\pm$1.15}

\newcommand{\CfDbTriadPm}{97.67$\pm$3.45}

\newcommand{\CfLrOAccPm}{99.89$\pm$0.17}

\newcommand{\CfLrOTriadPm}{99.67$\pm$0.50}

\newcommand{\CfNsAccPm}{99.89$\pm$0.17}

\newcommand{\CfNsTriadPm}{99.67$\pm$0.50}

\newcommand{\CfTxtAccMax}{71.89}
\newcommand{\CfTxtAccMin}{58.07}
\newcommand{\CfTxtLeakMax}{28.67}
\newcommand{\CfTxtLeakMin}{21.17}
\newcommand{\CfTxtTriadMax}{26.00}
\newcommand{\CfTxtTriadMin}{6.33}

\newcommand{\CfXgbOAccPm}{93.59$\pm$3.16}

\newcommand{\CfXgbOTriadPm}{82.11$\pm$8.05}

\newcommand{\ClauseMax}{38}

\newcommand{\ClauseMedian}{9}

\newcommand{\CmNsErrors}{16}

\newcommand{\CmNsHaramToHalal}{7}

\newcommand{\CmXgbHaramToHalal}{4}

\newcommand{\EcCarmN}{20}
\newcommand{\EcCarmNsAcc}{95.0}

\newcommand{\EcCarmXgbAcc}{100.0}

\newcommand{\EcGelN}{39}

\newcommand{\EcGlycN}{143}
\newcommand{\EcGlycNsAcc}{98.6}
\newcommand{\EcGlycNsFhlr}{1.2}

\newcommand{\EcGlycTxtFhlrMax}{10.7}
\newcommand{\EcGlycTxtFhlrMin}{7.1}

\newcommand{\EcGlyolDbFhlr}{18.8}

\newcommand{\EcGlyolN}{38}

\newcommand{\EcGlyolNsFhlr}{6.2}

\newcommand{\EcGlyolTxtFhlrMax}{25.0}
\newcommand{\EcGlyolTxtFhlrMin}{12.5}

\newcommand{\MsDbAccPm}{93.71$\pm$0.35}

\newcommand{\MsDbFhlrPm}{6.58$\pm$1.85}

\newcommand{\MsLrOAcc}{94.54}
\newcommand{\MsLrOAccPm}{94.54$\pm$0.85}

\newcommand{\MsLrOFhlr}{3.46}
\newcommand{\MsLrOFhlrPm}{3.46$\pm$0.31}

\newcommand{\MsLrOHaramPm}{2.62$\pm$1.25}

\newcommand{\MsNsAccPm}{97.04$\pm$0.34}
\newcommand{\MsNsAccSd}{0.34}

\newcommand{\MsNsAucPm}{99.84$\pm$0.08}

\newcommand{\MsNsFhlr}{4.68}
\newcommand{\MsNsFhlrPm}{4.68$\pm$1.16}

\newcommand{\MsNsHaramPm}{5.24$\pm$2.36}

\newcommand{\MsNsLat}{2.17}

\newcommand{\MsNsMfPm}{96.90$\pm$0.35}

\newcommand{\MsXgbHaramPm}{2.18$\pm$1.54}

\newcommand{\MsXgbLat}{0.07}

\newcommand{\MsXgbOAccPm}{96.25$\pm$1.24}
\newcommand{\MsXgbOAccSd}{1.24}

\newcommand{\MsXgbOFhlrPm}{3.81$\pm$0.83}

\newcommand{\MsXgbOHaramPm}{2.83$\pm$2.11}

\newcommand{\NCf}{540}
\newcommand{\NCfFamilies}{6}
\newcommand{\NDupText}{59}
\newcommand{\NEcode}{50}
\newcommand{\NEcodeHaram}{3}
\newcommand{\NEcodeMush}{29}
\newcommand{\NEcodeSafe}{18}
\newcommand{\NEscPat}{7}
\newcommand{\NHalal}{1{,}661}
\newcommand{\NHaram}{610}
\newcommand{\NHaramPat}{31}

\newcommand{\NItemsFmt}{3{,}199}
\newcommand{\NMush}{928}
\newcommand{\NMushPat}{32}
\newcommand{\NNegPat}{18}
\newcommand{\NQualPat}{18}
\newcommand{\NReal}{277}
\newcommand{\NRealHalal}{206}
\newcommand{\NRealHaram}{23}
\newcommand{\NRealMush}{48}

\newcommand{\NSynFmt}{2{,}922}

\newcommand{\NTest}{480}
\newcommand{\NTrain}{2239}
\newcommand{\NTriplets}{180}
\newcommand{\NVal}{480}

\newcommand{\PcHaramMax}{97.78}
\newcommand{\PcHaramMin}{92.40}

\newcommand{\PcMushDb}{92.20}
\newcommand{\PcMushGainXgb}{15.25}

\newcommand{\PcMushNs}{96.73}

\newcommand{\PcTxtMushMax}{81.48}
\newcommand{\PcTxtMushMin}{80.82}
\newcommand{\PctHalal}{51.9}
\newcommand{\PctHaram}{19.1}
\newcommand{\PctMush}{29.0}
\newcommand{\PctReal}{8.7}
\newcommand{\PctSyn}{91.3}
\newcommand{\PdNoAhrNoCaAcc}{+1.29}
\newcommand{\PdNoAhrNoCaAccHi}{+2.19}
\newcommand{\PdNoAhrNoCaAccLo}{+0.39}

\newcommand{\PdNsDbAcc}{+3.34}
\newcommand{\PdNsDbAccHi}{+3.75}
\newcommand{\PdNsDbAccLo}{+2.93}

\newcommand{\PdNsLrOAcc}{+2.50}

\newcommand{\PdNsLrOFhlr}{+1.21}

\newcommand{\PdNsLrOFhlrWins}{1}
\newcommand{\PdNsNoAhrAcc}{-0.50}
\newcommand{\PdNsNoAhrAccHi}{+0.07}
\newcommand{\PdNsNoAhrAccLo}{-1.06}

\newcommand{\PdNsXgbAcc}{+10.33}

\newcommand{\PdNsXgbOAcc}{+0.59}
\newcommand{\PdNsXgbOAccHi}{+2.28}
\newcommand{\PdNsXgbOAccLo}{-1.11}
\newcommand{\PdNsXgbOAccWins}{3}
\newcommand{\PdNsXgbOFhlr}{+1.21}
\newcommand{\PdNsXgbOFhlrHi}{+2.71}
\newcommand{\PdNsXgbOFhlrLo}{-0.28}
\newcommand{\PdNsXgbOFhlrWins}{1}
\newcommand{\ReLrOAccPm}{94.54$\pm$0.85}

\newcommand{\ReXgbOAccPm}{96.46$\pm$1.25}
\newcommand{\ReXgbOFhlrPm}{3.46$\pm$0.81}
\newcommand{\ReXgbOTriad}{81.89}

\newcommand{\SupHaram}{92}

\newcommand{\TrDbContPm}{1.26$\pm$1.43}

\newcommand{\TrDbGreenPm}{47.13$\pm$2.63}

\newcommand{\TrLrGreen}{22.25}

\newcommand{\TrLrManual}{77.75}

\newcommand{\TrLrOContPm}{0.96$\pm$0.75}

\newcommand{\TrLrOGreen}{42.96}
\newcommand{\TrLrOGreenPm}{42.96$\pm$0.86}

\newcommand{\TrNsCont}{0.76}
\newcommand{\TrNsContPm}{0.76$\pm$0.53}

\newcommand{\TrNsGreen}{49.33}
\newcommand{\TrNsGreenPm}{49.33$\pm$0.71}

\newcommand{\TrRfGreen}{18.63}

\newcommand{\TrRfManual}{81.37}

\newcommand{\TrXgbContPm}{4.11$\pm$1.68}

\newcommand{\TrXgbGreenPm}{35.96$\pm$2.59}

\newcommand{\TrXgbOCont}{1.46}
\newcommand{\TrXgbOContPm}{1.46$\pm$0.48}

\newcommand{\TrXgbOGreenPm}{48.54$\pm$0.91}

\newcommand{\TxtAccMax}{86.71}
\newcommand{\TxtAccMin}{86.25}

\newcommand{\TxtFhlrMax}{21.73}
\newcommand{\TxtFhlrMin}{13.42}

\newcommand{\TxtMfMax}{87.81}
\newcommand{\TxtMfMin}{86.27}

%% file: numbers_lto.tex
\newcommand{\LtoAllDbMushToHalal}{18.6}

\newcommand{\LtoAllLrOMushToHalal}{7.2}

\newcommand{\LtoAllNsMushToHalal}{11.1}

\newcommand{\LtoAllXgbOMushToHalal}{7.6}

\newcommand{\LtoDbAccPm}{86.85$\pm$14.89}

\newcommand{\LtoDbCfTriadPm}{79.63$\pm$38.17}

\newcommand{\LtoGelDbHalalToHaram}{73.8}

\newcommand{\LtoGelLrOHalalToHaram}{1.5}

\newcommand{\LtoGelNsHalalToHaram}{68.1}

\newcommand{\LtoGingDbMushToHalal}{98.5}

\newcommand{\LtoGingLrOMushToHalal}{4.5}

\newcommand{\LtoGingNsMushToHalal}{66.3}

\newcommand{\LtoLrAccPm}{78.35$\pm$11.92}

\newcommand{\LtoLrOAccMax}{94.40}
\newcommand{\LtoLrOAccMin}{87.13}
\newcommand{\LtoLrOAccPm}{91.11$\pm$2.88}
\newcommand{\LtoLrOAccSd}{2.88}

\newcommand{\LtoLrOCfTriadPm}{84.44$\pm$28.80}

\newcommand{\LtoNFolds}{10}
\newcommand{\LtoNSeeds}{3}

\newcommand{\LtoNoAhrAccPm}{91.88$\pm$11.03}

\newcommand{\LtoNsAccMax}{98.67}
\newcommand{\LtoNsAccMin}{64.64}
\newcommand{\LtoNsAccPm}{90.87$\pm$11.76}
\newcommand{\LtoNsAccSd}{11.76}

\newcommand{\LtoNsCfTriadPm}{92.41$\pm$13.08}

\newcommand{\LtoPdNsDbAcc}{+4.02}
\newcommand{\LtoPdNsDbAccHi}{+7.10}
\newcommand{\LtoPdNsDbAccLo}{+0.94}

\newcommand{\LtoPdNsDbFhlr}{-6.51}
\newcommand{\LtoPdNsDbFhlrHi}{-0.63}
\newcommand{\LtoPdNsDbFhlrLo}{-12.39}

\newcommand{\LtoPdNsLrOAcc}{-0.23}
\newcommand{\LtoPdNsLrOAccHi}{+7.87}
\newcommand{\LtoPdNsLrOAccLo}{-8.34}

\newcommand{\LtoPdNsNoAhrAcc}{-1.01}
\newcommand{\LtoPdNsNoAhrAccHi}{+0.24}
\newcommand{\LtoPdNsNoAhrAccLo}{-2.25}
\newcommand{\LtoPdNsNoAhrAccWins}{2}

\newcommand{\LtoPdNsXgbOAcc}{+4.18}
\newcommand{\LtoPdNsXgbOAccHi}{+13.63}
\newcommand{\LtoPdNsXgbOAccLo}{-5.27}

\newcommand{\LtoRfAccPm}{61.91$\pm$14.77}

\newcommand{\LtoTestMax}{329}
\newcommand{\LtoTestMin}{250}
\newcommand{\LtoTrainMean}{2470}
\newcommand{\LtoTxtAccMax}{78.35}
\newcommand{\LtoTxtAccMin}{61.91}

\newcommand{\LtoTxtCfTriadMax}{2.96}

\newcommand{\LtoValMean}{436}

\newcommand{\LtoXgbAccPm}{67.10$\pm$18.26}

\newcommand{\LtoXgbOAccPm}{86.70$\pm$5.50}

\newcommand{\LtoXgbOCfTriadPm}{52.96$\pm$47.65}

%% file: sections/intro.tex
\section{Introduction}
\label{sec:intro}

\IEEEPARstart{H}{alal} integrity of a food product is not decided once at the factory: it has to be preserved at every node where goods are stored, consolidated and cleared \cite{tieman2011application,tieman2012principles,soon2017halal,abtalib2015critical}.
We organise the discussion around three pillars: \emph{certification} by accredited bodies \cite{riaz2004halal,regenstein2003kosher,haleem2020halal}, \emph{traceability} of ingredients and batches \cite{rejeb2018halal,rejeb2021internet,tan2022blockchain,alourani2024blockchain}, and \emph{quality assurance}, which includes the segregation that prevents cross-contamination between Halal and Haram cargo \cite{tieman2013establishing,demirci2016positioning} and whose adoption by service providers has been reported to be low \cite{ngah2014adoption}.
Warehouse management systems, sensors and ledgers record what is known about a product \cite{ali2021sustainable,tan2022blockchain}; they do not decide what an ingredient declaration means.

Checking the declared ingredients against Halal rules is a document-based task, and undeclared ingredients and processing aids are a documented source of doubt \cite{almazeedi2013undeclared}.
Declarations are semi-structured: in our benchmark the median product has \ClauseMedian\ comma- or semicolon-delimited clauses (at most \ClauseMax), mixing culinary terms, chemical names and E-numbers listed in the Codex general standard for food additives \cite{codex2023gsfa}.
Some substances are unconditionally prohibited (e.g.\ pork fat, carmine E120) or permitted, but a large class of emulsifiers, humectants and enzymes can be manufactured from botanical, microbial or animal feedstock \cite{almazeedi2013undeclared,riaz2004halal,lubis2016from}; examples are mono- and diglycerides of fatty acids (E471, re-evaluated as a food additive by EFSA \cite{efsa2017re,efsa2021re}), glycerol (E422), gelatine (E441) and L-cysteine (E920).
When the declaration does not state an acceptable origin for such an additive, the product is \emph{Mushbooh} (doubtful) and is held for document verification before intake.
The qualifier is \emph{scoped}: in ``bakery shortening (certified plant-based palm oil), emulsifier: E471'' the plant-based qualifier belongs to the shortening, not to the emulsifier.

Automated screening of Halal status from packaging has been studied as label recognition with optical character recognition \cite{kartiwi2018mobile,tarannum2024halalcheck,arief2026halalbench}, detection in packaging images \cite{fadhilah2018nonhalal}, classification of ingredient lists \cite{bagundang2025ml,mulyana2025bilstm} and knowledge-graph completion for cosmetics \cite{hoang2025halal}.
To our reading of the abstracts, none of these studies evaluates whether a model respects the scope of an origin qualifier, that is, whether moving a qualifier from one clause to another changes the prediction as it should (Section~\ref{sec:related}).
A bag-of-words classifier discards clause boundaries by construction; a pre-trained transformer \cite{devlin2019bert,sanh2019distilbert} sees word order but is neither told where a clause ends nor what an E-number means.
The cost of an error is also asymmetric \cite{elkan2001foundations}: a \emph{false-Halal} decision (a Non-Halal or Mushbooh item predicted Halal) can contaminate a certified storage zone, whereas a \emph{false-quarantine} decision only delays a shipment.

This paper makes four contributions.
\textbf{(i)} A clause-scoped labelling pipeline, a \NItemsFmt-item benchmark and a \NCf-item (\NTriplets-triplet) counterfactual modifier-swap suite (Section~\ref{sec:data}); labels come from our rule base, not from a certification authority, and \PctSyn\,\% of the items are template-generated.
\textbf{(ii)} \nesy{}, a dual-stream network in which a bank of four queries derived from a 16-dimensional rule-based symbolic vector attends over DistilBERT token states, followed by gated fusion (Section~\ref{sec:method}).
\textbf{(iii)} A five-seed evaluation against strong baselines that receive the \emph{same} symbolic vector, with seed-paired differences, and a leave-one-template-family-out evaluation on ten held-out generator families (Section~\ref{sec:results}); it shows that the symbolic features, not the neural architecture, carry most of the gain, and that an asymmetric risk loss designed to penalise false-Halal decisions did not help.
\textbf{(iv)} A three-lane receiving-dock triage simulation that compares all models, including the ontology-augmented baselines, at a fixed confidence threshold.
We make no claim of priority; related work is discussed in Section~\ref{sec:related}.
Because the labels and the symbolic features are produced by the same rule engine, high accuracy here measures fidelity to a documented rule policy and not Halal certification; Section~\ref{sec:discussion} lists the resulting limits.

%% file: sections/related.tex
\section{Related Work}
\label{sec:related}

\paragraph*{Halal supply chain integrity}
Early work described the basic requirements of Halal supply chains as avoiding direct contact with Haram and addressing the risk of contamination \cite{tieman2011application}, proposed a Halal Supply Chain Model with control and assurance activities in transportation, warehousing and terminal operations \cite{tieman2012principles}, and set out the principles of Halal logistics, including how to define and avoid cross-contamination between Halal and Haram \cite{tieman2013establishing}.
Later studies examined Halal integrity along the food chain \cite{soon2017halal}, critical success factors, which include Halal certification and traceability \cite{abtalib2015critical}, the link between certification and logistics performance \cite{abtalib2016halal}, drivers of a Halal orientation strategy in firms \cite{zailani2015drivers}, barriers to the adoption of certification \cite{haleem2020halal} and definitions of Halal supply chain management \cite{khan2018halal}; in interviews with three Malaysian service providers, the adoption of Halal transportation and warehousing was reported to be low and to face barriers \cite{ngah2014adoption}.
For traceability, Halal rules have been embedded into HACCP, blockchain and Internet-of-Things systems \cite{rejeb2018halal,rejeb2021internet,ali2021sustainable,tan2022blockchain}, and a blockchain and AI system has been tested in a supermarket \cite{alourani2024blockchain}.
Such systems record the status of a product once it is known.
Deciding the status of an ingredient declaration is upstream of them, and the authentication literature documents several sources of doubt: undeclared processing aids \cite{almazeedi2013undeclared}, pig derivatives such as lard, pork and gelatin, for which analytical detection methods exist \cite{rohman2012analysis}, and authenticity issues in meat products \cite{nakyinsige2012halal}; see also \cite{lubis2016from,bonne2008determinants}.

\paragraph*{Food computing and Open Food Facts}
Food computing, surveyed in \cite{min2019survey}, works with food images, recipes and other food data.
Related text-based tasks include macronutrient prediction from short descriptions of foods \cite{ispirova2020pnut} and the prediction of the degree of food processing \cite{menichetti2023machine}; Recipe1M+ links over one million recipes with food images \cite{marin2019recipe1m}.
Open Food Facts \cite{openfoodfacts} is a crowd-sourced product database that we use as the source of real declarations.
None of the cited tasks concerns the permissibility of an ingredient.

\paragraph*{Automated Halal screening}
Screening from packaging has been built as an OCR pipeline on a smartphone \cite{kartiwi2018mobile}, as character-level CNN recognition of the ingredient image followed by comparison with a list of non-Halal compositions \cite{fadhilah2018nonhalal}, and as YOLOv5 detection plus OCR followed by machine-learning, neural and rule-based classification of Halal or Haram status \cite{tarannum2024halalcheck}; HalalBench benchmarks OCR engines on ingredient labels \cite{arief2026halalbench}.
On the text side, Naive Bayes, logistic regression, SVM and random forests were compared on a Halal/Haram ingredient data set, where logistic regression and SVM reached 99\,\% accuracy \cite{bagundang2025ml}, and a bidirectional LSTM \cite{hochreiter1997long,schuster1997bidirectional} with Word2Vec and GloVe embeddings \cite{mikolov2013efficient,pennington2014glove} reached 99.75\,\% on 3,979 samples \cite{mulyana2025bilstm}.
For cosmetics, a knowledge graph of products and ingredients was used to predict Halal status \cite{hoang2025halal}.
Table~\ref{tab:related} summarises what these studies state in their abstracts.
To our reading, none reports a test in which the clause that carries an origin qualifier is varied while the rest of the declaration is held fixed; this is a statement about what we could read, not proof of absence.
Binary Halal/Haram ingredient classification is evidently easy for linear models \cite{bagundang2025ml}; the three-way task with scoped qualifiers is what our benchmark adds.

\paragraph*{Neuro-symbolic learning and knowledge-enhanced encoders}
Neuro-symbolic AI seeks to combine neural representations with explicit domain knowledge \cite{garcez2023neurosymbolic,besold2021neural,marcus2020next}.
For language encoders, K-BERT injects knowledge-graph triples into the sentence and limits their influence with soft positions and a visible matrix \cite{liu2020kbert}, ERNIE trains on text and knowledge graphs jointly \cite{zhang2019ernie}, and KnowBert updates contextual word representations by word-to-entity attention \cite{peters2019knowledge}.
Our encoder is a pre-trained DistilBERT \cite{sanh2019distilbert} (cf.\ BERT and RoBERTa \cite{devlin2019bert,liu2019roberta}) that is not modified; the symbolic knowledge enters through a small bank of query vectors that attend over its token states with multi-head attention \cite{vaswani2017attention}, in the spirit of the asymmetric attention of the Perceiver, where a small latent array attends over a large input \cite{jaegle2021perceiver}.

\paragraph*{Rule-derived labels, shortcuts and counterfactual evaluation}
When labels come from heuristics, weak-supervision systems treat them as noisy labelling functions \cite{ratner2017snorkel}; here the rules are the specification and we say so.
Models that score well on a benchmark can rely on shortcuts that fail elsewhere \cite{geirhos2020shortcut,mccoy2019right}, and crowd-written data can leave annotation artifacts \cite{gururangan2018annotation}; counterfactual contrast sets, counterfactually augmented data and behavioural tests probe such failures by changing only the factor of interest \cite{kaushik2020learning,gardner2020evaluating,ribeiro2020beyond}.
Our counterfactual suite follows this logic for the scope of an origin qualifier.

\paragraph*{Asymmetric costs and selective prediction}
Cost-sensitive learning, class re-weighting and loss modulation address unequal error costs and class imbalance \cite{elkan2001foundations,chawla2002smote,lin2017focal}.
Rejecting uncertain inputs has a long history as the reject option \cite{chow1970optimum,geifman2017selective}, and the maximum softmax probability is a common baseline for deciding when to trust a prediction \cite{hendrycks2017baseline}, although modern networks are over-confident \cite{guo2017calibration}.
Our three-lane receiving simulation is a reject-option rule with a fixed, untuned threshold on that confidence.

\begin{table*}[!t]
\caption{Positioning against the closest automated Halal screening studies. Entries are taken from the abstracts of the cited works; ``not stated'' means that the abstract does not mention the item, not that the study lacks it.}
\label{tab:related}
\centering\footnotesize
\newcolumntype{L}[1]{>{\raggedright\arraybackslash}p{#1}}
\begin{tabular}{@{}L{3.0cm}L{3.0cm}L{5.0cm}L{2.8cm}L{2.5cm}@{}}
\toprule
Work & Input & Method (as stated) & Output labels & Qualifier scope tested \\
\midrule
Kartiwi \emph{et al.} \cite{kartiwi2018mobile} & photo of ingredient list & OCR, then flag unlawful items & lawful / unlawful items & not stated \\
Fadhilah \emph{et al.} \cite{fadhilah2018nonhalal} & photo of packaging & character CNN, word list of non-Halal compositions & non-Halal detection & not stated \\
Tarannum \emph{et al.} \cite{tarannum2024halalcheck} & photo of packaging & YOLOv5 + OCR, then ML, neural and rule-based classifiers & Halal / Haram & not stated \\
Bagundang and Gallo \cite{bagundang2025ml} & ingredient list (text) & Naive Bayes, LR, SVM, random forest & Halal / Haram & not stated \\
Mulyana \emph{et al.} \cite{mulyana2025bilstm} & ingredient list (text) & BiLSTM, Word2Vec / GloVe & Halal / Haram & not stated \\
Hoang \emph{et al.} \cite{hoang2025halal} & cosmetics-ingredient graph & knowledge-graph completion & Halal status (cosmetics) & not applicable \\
\textbf{This study} & ingredient text + additive tags & DistilBERT + symbolic cross-attention, strong baselines with the same vector & Halal / Non-Halal / Mushbooh & yes, 180 counterfactual triplets \\
\bottomrule
\end{tabular}
\end{table*}

%% file: sections/data.tex
\section{Problem and Benchmark}
\label{sec:data}

\subsection{Cross-clause modifier confounding}
\begin{figure*}[!t]
\centering
\includegraphics[width=\textwidth]{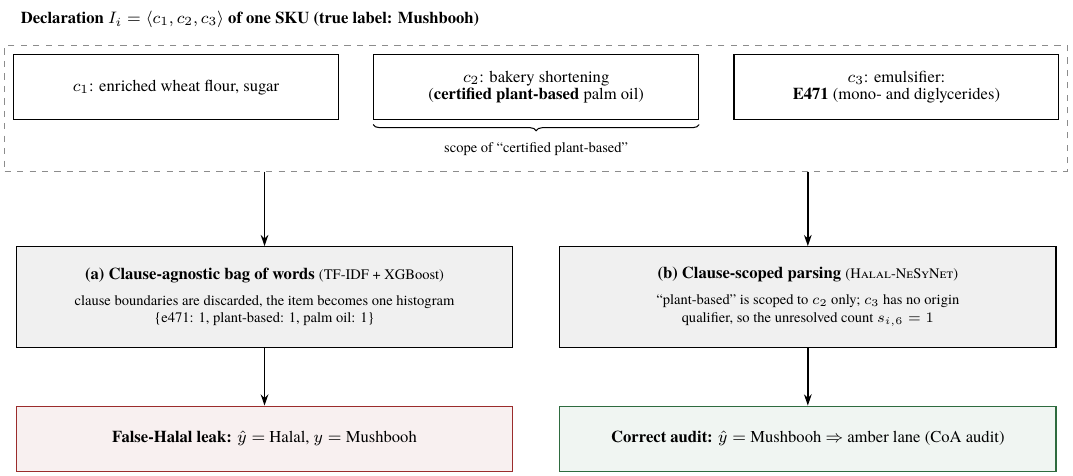}
\caption{Cross-clause modifier confounding. A plant-based qualifier on a benign shortening clause ($c_2$) coexists with an unqualified E471 emulsifier in $c_3$. A bag of words merges the clauses and predicts Halal, a false-Halal leak; a clause-scoped parser keeps the qualifier with $c_2$, flags $c_3$ as unresolved ($s_{i,6}=1$, Table~\ref{tab:sym}) and routes the item to the amber lane.}
\label{fig:confound}
\end{figure*}

A product $i$ is described by $x_i=(I_i,A_i,L_i)$: the ingredient declaration $I_i$, the structured additive tags $A_i$ and the label tags $L_i$.
A parenthesis-balanced parser splits $I_i$ at top-level commas and semicolons into an ordered sequence of $m_i$ clauses, $I_i=\langle c_{i,1},\dots,c_{i,m_i}\rangle$.
Let clause $c_{i,k}$ contain a dual-origin additive $e$ (for instance E471 or E422).
Under clause-scoped auditing, $e$ is resolved to Halal only if a botanical, microbial or certified-Halal qualifier modifies it \emph{within the same clause}, and to Non-Halal if an animal or porcine qualifier does; otherwise it stays Mushbooh.
When the plant-based qualifier sits instead on a different, benign clause $c_{i,j}$ ($j\neq k$), any classifier whose features are invariant to clause permutation, such as a bag of words, cannot tell the two situations apart.
Figure~\ref{fig:confound} shows the failure: the plant-based qualifier of the shortening clears the unqualified emulsifier E471.

\subsection{Data sources}
\textbf{Real records.} We queried the Open Food Facts REST API \cite{openfoodfacts} for products carrying the \texttt{en:halal} tag and for confectionery and biscuit categories, keeping records with non-empty English \texttt{ingredients\_text}.
After cleaning this gave \NReal\ real products (\NRealHalal\ Halal, \NRealMush\ Mushbooh, \NRealHaram\ Non-Halal under our rules).

\textbf{Template-generated items.} Crowd-sourced databases hold few Mushbooh and Non-Halal examples with systematically varied qualifier placement, so we generated \NSynFmt\ further formulations to reach a target of about 3,200 items.
A generated item combines a grain base, two benign additive clauses and either one of eight dual-origin additive templates (E471, E472e, E441, E422, rennet and chymosin, E920, E433, E631) or, with probability 0.18, one of two negation templates (ginger beer and imitation bacon bits, wine vinegar).
The additive clause and a benign carrier clause share the bridge phrase ``specification:''.
The additive clause carries the wording that sets the class (plant origin for Halal, neutral commercial wording for Mushbooh, animal origin for Non-Halal); in 45\,\% of the additive-template items the benign carrier clause additionally carries the contrasting wording, for example a plant-origin qualifier next to an unqualified additive.
\emph{By construction}, therefore, the unigram and bigram content of such an item is nearly unchanged when the qualifier moves between clauses; a bag of words is expected to fail on them, and the experiment asks which models can use clause scope.
Overall \PctReal\,\% of the \NItemsFmt\ items are real and \PctSyn\,\% are template-generated (Table~\ref{tab:data}); the class totals are \NHalal\ Halal (\PctHalal\,\%), \NMush\ Mushbooh (\PctMush\,\%) and \NHaram\ Non-Halal (\PctHaram\,\%).
Because the generated items come from a small set of templates, \NDupText\ declaration texts occur more than once.

\begin{table}[!t]
\caption{Benchmark composition by source and class (labels from the rule base).}
\label{tab:data}
\centering\footnotesize
\input{tables/tab_data}
\end{table}

\subsection{Rule engine, labels and the symbolic vector}
\label{sec:labels}
Text is lower-cased, HTML and Unicode artifacts are stripped, and E-code orthography (\texttt{E-471}, \texttt{E 471}, \texttt{INS 471}) is normalised (\texttt{e471}).
The rule base has \NEcode\ E-codes (\NEcodeHaram\ Haram, \NEcodeMush\ Mushbooh, \NEcodeSafe\ Halal-safe) and four pattern lists: \NHaramPat\ Haram patterns (pork products, porcine, alcoholic drinks, carmine and cochineal, shellac, bone phosphate), \NNegPat\ negation patterns that neutralise them (ginger beer, non-alcoholic, wine vinegar, imitation bacon, \dots), \NMushPat\ dual-origin patterns (gelatine, mono- and diglycerides, glycerol, rennet, whey, L-cysteine, polysorbates, tallow, \dots) and \NQualPat\ Halal-qualifier patterns (plant-based, vegetable origin, microbial, vegan, halal, synthetic, \dots) with \NEscPat\ escalators (porcine, pork, lard, non-halal, insect, \dots).
It was compiled from the Codex general standard for food additives \cite{codex2023gsfa}, EFSA evaluations \cite{efsa2017re,efsa2021re} and the Halal food science literature \cite{riaz2004halal,almazeedi2013undeclared,nakyinsige2012halal,rohman2012analysis}; it has not been audited by a Halal certification body and represents one conservative policy.
Each clause is examined in turn and the decision hierarchy is:
\begin{enumerate}
\item \emph{Non-Halal} ($y=1$) if a Haram E-code is present, a clause has an un-negated Haram pattern, or a dual-origin clause carries an escalator;
\item otherwise \emph{Halal} ($y=0$) if the product carries an \texttt{en:halal}, \texttt{en:vegan} or \texttt{en:vegetarian} tag;
\item otherwise \emph{Mushbooh} ($y=2$) if some clause contains a dual-origin E-code or pattern without a Halal qualifier \emph{in the same clause} (or a Mushbooh E-code appears only in the tags);
\item otherwise \emph{Halal}.
\end{enumerate}
The same parser outputs a 16-dimensional symbolic vector $\sv_i\in\mathbb{R}^{16}$ of raw counts and flags (Table~\ref{tab:sym}) computed from the text and tags alone, so it is available at inference without labels.

\begin{table}[!t]
\caption{The 16-dimensional symbolic vector $\sv_i$ produced by the clause parser (raw counts unless stated; $m_i$ is the number of clauses).}
\label{tab:sym}
\centering\footnotesize
\renewcommand{\arraystretch}{1.05}
\begin{tabular}{@{}p{1.5cm}>{\raggedright\arraybackslash}p{6.3cm}@{}}
\toprule
Index & Feature \\
\midrule
$s_0,s_1$ & explicit \texttt{en:halal} tag; \texttt{en:vegan} or \texttt{en:vegetarian} tag (0/1) \\
$s_2$ & number of Haram E-codes (text or tags) \\
$s_3$ & clauses with an un-negated Haram pattern or E-code \\
$s_4$ & clauses with a negated Haram phrase (e.g.\ ginger beer) \\
$s_5$ & number of Mushbooh E-codes \\
$s_6$ & dual-origin clauses \emph{without} a local qualifier (unresolved) \\
$s_7$ & dual-origin clauses resolved by a local plant, microbial or Halal qualifier \\
$s_8$ & dual-origin clauses escalated by an animal or porcine modifier \\
$s_9$ & number of Halal-safe E-codes \\
$s_{10}$--$s_{14}$ & flags: carmine or E120; E471/E472 or diglycerides; glycerol or E422; gelatine or E441; whey, rennet, pepsin or lipase (0/1) \\
$s_{15}$ & $\min(1,\,m_i/20)$ \\
\bottomrule
\end{tabular}
\end{table}

\subsection{Input perturbations and splits}
\label{sec:perturb}
Labels and $\sv_i$ are computed from the declaration \emph{before} the following perturbations, which mimic OCR noise, supplier vocabulary drift and parser errors on non-counterfactual items: (i) in 35\,\% of items the bridge phrase ``specification:'' is replaced by ``batch note --'' in the text; (ii) in 22\,\% explicit modifier phrases in the text are replaced by unseen technical synonyms (``certified plant-based'' $\to$ ``phytolipid ester'', ``porcine lard fat'' $\to$ ``suidae adipose'', ``halal certified bovine'' $\to$ ``ruminant certified'', \dots); (iii) in 3.6\,\% the parenthetical qualifier becomes ``origin unspecified'' and all entries of $\sv_i$ are replaced by uniform noise in $[0,0.5]$ while the label is kept, i.e.\ about 3.6\,\% label noise; and (iv) in the remaining items each of $s_3,s_6,s_7,s_8$ is changed by $\pm1$ with probability 0.14 to mimic parser errors.
Perturbations (i) and (ii) change what the text-based models read but not $\sv_i$: the symbolic vector keeps the qualifier information that the perturbed text hides.
This favours every model that receives $\sv_i$ and is a limit of the benchmark (Section~\ref{sec:discussion}).
The \NItemsFmt\ items are split by stratified 70/15/15 sampling into \NTrain\ training, \NVal\ validation and \NTest\ test items; every experiment is repeated for five seeds $\{42,\dots,46\}$ that change the split, the weight initialisation and the tree sampling.

\subsection{Counterfactual modifier-swap suite}
The suite has \NCf\ items in \NTriplets\ triplets over \NCfFamilies\ additive families (E471, E441, E422, E433 with E120, rennet in dairy, E920).
Within a triplet the declaration is fixed and only the clause-local modifier on the target additive changes: unverified origin (Mushbooh), plant or microbial origin (Halal), or porcine or animal origin (Non-Halal), while a benign clause may carry a contrasting qualifier.
In every fourth triplet the qualifier words are replaced by chemical synonyms, and in a few items the qualifier is truncated to ``origin unspecified'' with a noisy vector, as in the main data.
No suite text occurs in the main benchmark, but the suite shares templates with it, so it measures sensitivity to local qualifier swaps and not out-of-domain generalisation.

%% file: tables/tab_data.tex
\begin{tabular}{@{}lrrrr@{}}
\toprule
Source & Halal & Mushbooh & Non-Halal & Total \\
\midrule
Open Food Facts records & 206 & 48 & 23 & 277 \\
Template-generated items & 1{,}455 & 880 & 587 & 2{,}922 \\
\midrule
All items & 1{,}661 & 928 & 610 & 3{,}199 \\
\bottomrule
\end{tabular}

%% file: sections/method.tex
\section{Halal-NeSyNet and Experimental Protocol}
\label{sec:method}

\begin{figure*}[!t]
\centering
\includegraphics[width=\textwidth]{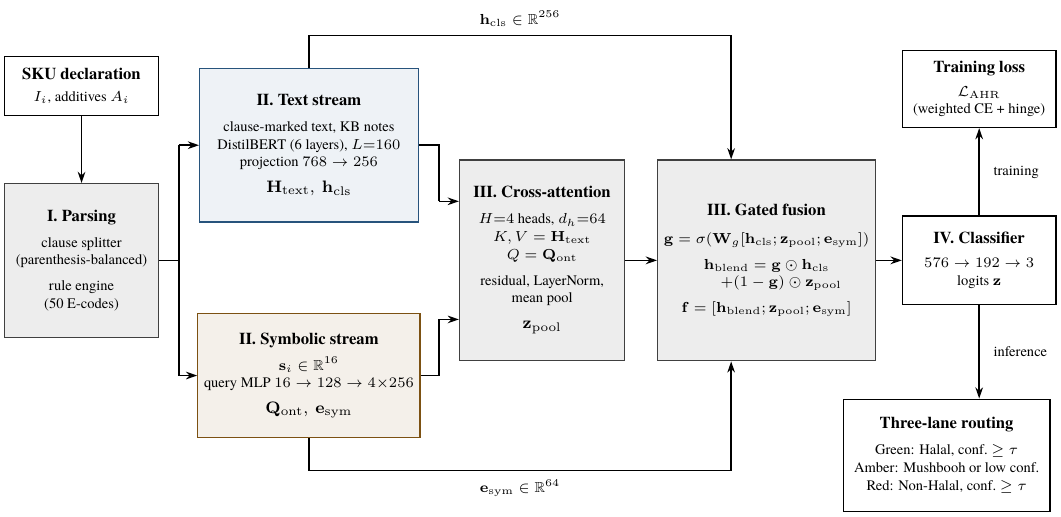}
\caption{Architecture of \nesy{}. (I) A parenthesis-balanced clause splitter and the rule engine turn a SKU declaration into clause-marked text and a 16-dimensional symbolic vector $\sv_i$. (II) A DistilBERT text stream yields token states $\mathbf{H}_\mathrm{text}$ and $\mathbf{h}_\mathrm{cls}$; a query MLP turns $\sv_i$ into four queries $\mathbf{Q}_\mathrm{ont}$ and a direct embedding $\mathbf{e}_\mathrm{sym}$. (III) The queries attend over the token states, and a sigmoid gate blends $\mathbf{h}_\mathrm{cls}$ with the pooled attention output. (IV) A classifier produces three logits, trained with $\ahr$; at inference the maximum softmax confidence, compared with the threshold $\tau$, routes the item to the green, amber or red lane.}
\label{fig:arch}
\end{figure*}

Figure~\ref{fig:arch} shows the model.
Unlike a single-stream text classifier, \nesy{} reads the clause-marked text and the symbolic vector in parallel and binds them with cross-attention before a gated fusion layer.

\subsection{Dual-stream encoding, cross-attention and gated fusion}
\textbf{Text stream.}
The segmented clauses and the recognised E-code group names of the rule base are formatted as
\begin{center}
\texttt{clauses: [c1] $c_{i,1}$ ; [c2] $c_{i,2}$ \dots{} | additives: $A_i$ | kb: e471:dual\_fat\_emulsifier \dots}
\end{center}
truncated or padded to $L=160$ WordPiece tokens and encoded by \texttt{distilbert-base-uncased} \cite{sanh2019distilbert} into $\mathbf{U}_i\in\mathbb{R}^{L\times768}$.
A linear projection with layer normalisation \cite{ba2016layer} and GELU gives
\begin{align}
\mathbf{H}_\mathrm{text}&=\mathrm{GELU}\big(\mathrm{LN}(\mathbf{U}_i\mathbf{W}_\mathrm{proj}+\mathbf{b}_\mathrm{proj})\big)\in\mathbb{R}^{L\times256},\\
\mathbf{h}_\mathrm{cls}&=\mathbf{H}_{\mathrm{text},0}.
\end{align}

\textbf{Symbolic stream.}
A two-layer MLP maps $\sv_i$ to $K=4$ query vectors and, by a separate projection, to a direct embedding:
\begin{align}
\mathbf{u}&=\mathrm{Drop}\big(\mathrm{GELU}(\mathrm{LN}(\mathbf{W}_{q,1}\sv_i))\big),\\
\mathbf{Q}_\mathrm{ont}&=\mathrm{LN}\big(\mathrm{Reshape}_{4\times256}(\mathrm{GELU}(\mathbf{W}_{q,2}\mathbf{u}))\big),\\
\mathbf{e}_\mathrm{sym}&=\mathrm{GELU}\big(\mathrm{LN}(\mathbf{W}_e\sv_i)\big)\in\mathbb{R}^{64}.
\end{align}
The four queries are not given distinct roles by any loss; any specialisation would have to emerge from training.

\textbf{Cross-attention and gated fusion.}
The queries attend over the token states (keys and values) with $H=4$ heads of size 64 \cite{vaswani2017attention}, padded positions masked; a residual connection \cite{he2016deep} and layer normalisation follow, and the four outputs are averaged:
\begin{align}
\mathbf{Z}&=\mathrm{LN}\big(\mathbf{Q}_\mathrm{ont}+\mathrm{MHA}(\mathbf{Q}_\mathrm{ont},\mathbf{H}_\mathrm{text},\mathbf{H}_\mathrm{text})\big),\\
\mathbf{z}_\mathrm{pool}&=\tfrac14\textstyle\sum_{k=1}^{4}\mathbf{Z}_k .
\end{align}
A sigmoid gate blends the contextual and the cross-attended representation,
\begin{align}
\mathbf{g}&=\sigma\big(\mathbf{W}_g[\mathbf{h}_\mathrm{cls};\mathbf{z}_\mathrm{pool};\mathbf{e}_\mathrm{sym}]\big),\\
\mathbf{h}_\mathrm{blend}&=\mathbf{g}\odot\mathbf{h}_\mathrm{cls}+(1-\mathbf{g})\odot\mathbf{z}_\mathrm{pool},
\end{align}
and $\mathbf{f}=[\mathbf{h}_\mathrm{blend};\mathbf{z}_\mathrm{pool};\mathbf{e}_\mathrm{sym}]\in\mathbb{R}^{576}$ feeds a classifier ($576\to192\to3$, layer norm, GELU, dropout 0.2).
The design resembles knowledge-enhanced encoders in that a pre-trained encoder is combined with structured knowledge \cite{liu2020kbert,zhang2019ernie,peters2019knowledge}, but the encoder is left unchanged and the knowledge enters as a small set of queries, as in latent-array attention \cite{jaegle2021perceiver}.

\subsection{Asymmetric Halal risk loss}
A false-Halal decision is costlier than a false quarantine \cite{elkan2001foundations}.
We add to the class-weighted cross-entropy $\mathcal{L}_\mathrm{WCE}$ (weights inversely proportional to the training class frequencies, a standard remedy for imbalance \cite{chawla2002smote,lin2017focal}) a quadratic hinge penalty whenever the logit of Halal approaches that of the true class of a Non-Halal or Mushbooh item:
\begin{equation}
\ahr(\mathbf{z},y)=\mathcal{L}_\mathrm{WCE}(\mathbf{z},y)+\alpha_y\max\big(0,\;z_0-z_y+\delta\big)^2,
\end{equation}
with $\delta=0.5$, $\alpha_0=0$, $\alpha_1=0.65$ (Non-Halal) and $\alpha_2=0.40$ (Mushbooh); the penalty is averaged over the batch.
These values were set before the runs and not tuned on the validation set.

\subsection{Baselines, training and software}
All models see the same stratified split.
\textbf{Text-only baselines:} TF-IDF on word 1--2-grams (12,000 features, sublinear term frequency) with balanced logistic regression (LR, $C=1.5$), random forest (RF, 200 trees, depth 22) \cite{breiman2001random} and XGBoost (200 trees, depth 5, learning rate 0.08, subsampling 0.85) \cite{chen2016xgboost}.
\textbf{Ontology-augmented baselines:} the same LR and XGBoost on the concatenation of the TF-IDF vector and $\sv_i$; these isolate how much of \nesy{}'s gain comes from the symbolic vector rather than from the network.
\textbf{Vanilla DistilBERT:} fine-tuned on the plain text with $L=96$, AdamW \cite{loshchilov2019decoupled} (learning rate $2.5\times10^{-5}$, weight decay 0.01, 10\,\% warm-up), batch size 16, three epochs and class-weighted cross-entropy.
\textbf{\nesy{}:} $L=160$, learning rate $3\times10^{-5}$, otherwise the same schedule; for both neural models the epoch with the best validation macro-F1 is kept.
The two ablations are trained like \nesy{}: \emph{w/o $\ahr$} uses weighted cross-entropy; \emph{w/o cross-attention} removes the attention branch and, in the released code, is also trained without $\ahr$, on the plain text with $L=112$ (Table~\ref{tab:ablation} marks this).
Implementation uses scikit-learn \cite{pedregosa2011scikit}, XGBoost, PyTorch \cite{paszke2019pytorch} and Transformers \cite{wolf2020transformers}; the neural runs used a single NVIDIA RTX 5060 GPU.
The per-seed results of the ontology-augmented baselines were not stored by the original run; we re-ran them (XGBoost 3.2.0, scikit-learn 1.9.0) to obtain paired values (Section~\ref{sec:results}).

\subsection{Metrics and statistics}
We report accuracy, macro-F1 and one-vs-rest macro ROC-AUC, and two supply-chain safety metrics: the \emph{false-Halal leakage rate} (FHLR), the share of true Non-Halal or Mushbooh test items predicted Halal, and the \emph{Haram$\to$Halal rate}, the share of true Non-Halal items predicted Halal.
On the counterfactual suite we report the item accuracy, the \emph{strict triad consistency} (all three members of a triplet correct) and the counterfactual FHLR.
For triage (Section~\ref{sec:triage}) an item is routed to the \emph{green lane} if the predicted class is Halal with softmax confidence $\geq\tau=0.85$, to the \emph{red lane} if it is predicted Non-Halal with confidence $\geq\tau$, and to the \emph{amber lane} (certificate-of-analysis audit) otherwise; the contamination rate is the share of green-lane items that are not Halal.
The confidence is the maximum softmax probability \cite{hendrycks2017baseline}, not calibrated \cite{guo2017calibration}, and $\tau$ was fixed a priori.
Values are means and sample standard deviations over five seeds.
Because five seeds are few, we also report seed-paired differences with $t$-based 95\,\% intervals (4 degrees of freedom) and the number of seeds in which a model is better; no correction for multiple comparisons is applied, and the intervals are descriptive.
Inference latency is in milliseconds per item (single-threaded CPU for the TF-IDF models, GPU with batch size 16 for the transformers) and is not comparable across families.

%% file: sections/results.tex
\section{Results}
\label{sec:results}

All numbers are means and sample standard deviations over five seeds on the held-out test split ($N=\NTest$ per run) unless a single seed is stated; every value is generated from the result files.

\subsection{Main benchmark}
\begin{table*}[!t]
\caption{Five-seed test results (mean $\pm$ SD). FHLR: share of true Non-Halal or Mushbooh items predicted Halal; Haram$\to$Halal: share of true Non-Halal items predicted Halal (lower is better for both). $\mathbf{s}$: the 16-dimensional symbolic vector. Best mean per column in bold. The two ``$+\,\mathbf{s}$'' rows are the values of the original run.}
\label{tab:main}
\centering\footnotesize
\input{tables/tab_main}
\end{table*}
Table~\ref{tab:main} gives the main results.
The three text-only TF-IDF baselines reach \TxtAccMin--\TxtAccMax\,\% accuracy and \TxtMfMin--\TxtMfMax\,\% macro-F1, with an FHLR of \TxtFhlrMin--\TxtFhlrMax\,\%.
Vanilla DistilBERT raises the accuracy to \MsDbAccPm\,\% and roughly halves the FHLR to \MsDbFhlrPm\,\%.
Supplying the symbolic vector to the classical models changes the picture more than changing the model family: logistic regression with $\mathbf{s}$ reaches \MsLrOAccPm\,\% (FHLR \MsLrOFhlrPm\,\%) and XGBoost with $\mathbf{s}$ \MsXgbOAccPm\,\% (FHLR \MsXgbOFhlrPm\,\%).
\nesy{} has the highest mean accuracy (\MsNsAccPm\,\%), macro-F1 (\MsNsMfPm\,\%) and ROC-AUC (\MsNsAucPm\,\%), but its FHLR (\MsNsFhlrPm\,\%) is higher than that of both models with $\mathbf{s}$, and its Haram$\to$Halal rate (\MsNsHaramPm\,\%) is higher than that of XGBoost on text alone (\MsXgbHaramPm\,\%) as well as of the two models with $\mathbf{s}$ (\MsLrOHaramPm\,\% and \MsXgbOHaramPm\,\%).
Inference takes \MsNsLat\,ms per item on a GPU for \nesy{} and \MsXgbLat\,ms on a CPU for text-only XGBoost, so the latencies are not comparable.

\begin{figure*}[!t]
\centering
\includegraphics[width=0.94\textwidth]{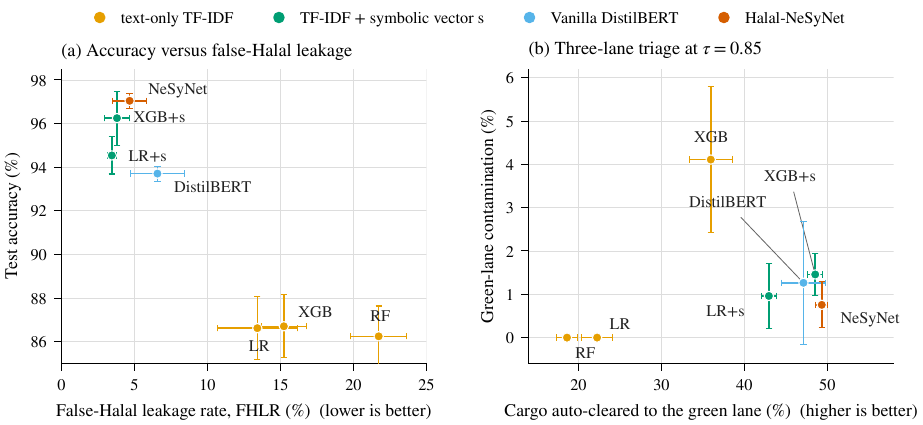}
\caption{Trade-offs (mean $\pm$ SD over five seeds). (a) Accuracy against false-Halal leakage; the best models sit at the upper left. (b) Share of cargo auto-cleared to the green lane against the contamination of that lane at $\tau=0.85$; the best models sit at the lower right. The models with the symbolic vector $\mathbf{s}$ and \nesy{} are close in both panels.}
\label{fig:tradeoff}
\end{figure*}

\subsection{Per-class performance}
\begin{figure*}[!t]
\centering
\includegraphics[width=0.96\textwidth]{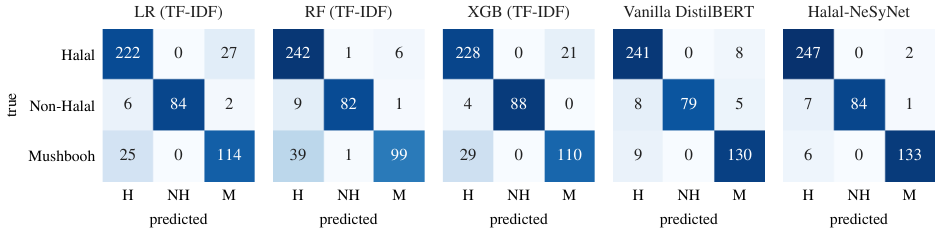}
\caption{Confusion matrices on the seed-42 test split ($N=\NTest$; H: Halal, NH: Non-Halal, M: Mushbooh; shading is the row-normalised share).}
\label{fig:cm}
\end{figure*}

\begin{table}[!t]
\caption{Per-class F1 (\%) on the test split of one run (seed 42); support is the number of test items.}
\label{tab:perclass}
\centering\footnotesize
\setlength{\tabcolsep}{4pt}
\input{tables/tab_perclass}
\end{table}
Table~\ref{tab:perclass} and Fig.~\ref{fig:cm} show the seed-42 split.
All five models identify the explicit Non-Halal items well (F1 of \PcHaramMin--\PcHaramMax\,\%), because overt Haram terms give strong lexical signals once culinary negations are filtered.
The classes separate the models on Mushbooh: the text-only models reach \PcTxtMushMin--\PcTxtMushMax\,\%, Vanilla DistilBERT \PcMushDb\,\% and \nesy{} \PcMushNs\,\%, \PcMushGainXgb\ points above text-only XGBoost.
The most serious error is a true Non-Halal item predicted Halal: \CmNsHaramToHalal\ of \SupHaram\ items for \nesy{} and \CmXgbHaramToHalal\ for text-only XGBoost; \nesy{} makes \CmNsErrors\ errors in total.

\subsection{Is the network needed? Seed-paired comparison}
\begin{figure}[!t]
\centering
\includegraphics[width=\columnwidth]{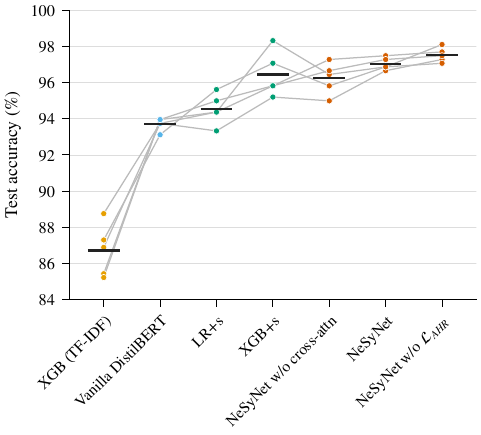}
\caption{Test accuracy of every seed (dots) and the five-seed mean (black bars); grey lines join the same seed. The models with $\mathbf{s}$ and \nesy{} overlap; XGB+s uses the re-run.}
\label{fig:paired}
\end{figure}

\begin{table*}[!t]
\caption{Seed-paired differences over the five seeds (first model minus second; mean, $t$-based 95\,\% interval, number of seeds in which the first model is better). Rows with $\mathbf{s}$ use the re-run of the ontology-augmented baselines. $^\dagger$The second model is also trained without $\ahr$, on the plain text with $L=112$.}
\label{tab:paired}
\centering\footnotesize
\input{tables/tab_paired}
\end{table*}
Table~\ref{tab:paired} and Fig.~\ref{fig:paired} compare the models seed by seed.
\nesy{} is more accurate than Vanilla DistilBERT by $\PdNsDbAcc$ points (95\,\% interval [$\PdNsDbAccLo$, $\PdNsDbAccHi$]) in all five seeds, and than text-only XGBoost by $\PdNsXgbAcc$ points in all five seeds.
Against baselines that receive the same symbolic vector the advantage is small or absent: relative to XGBoost with $\mathbf{s}$ it is $\PdNsXgbOAcc$ points ([$\PdNsXgbOAccLo$, $\PdNsXgbOAccHi$]), better in \PdNsXgbOAccWins\ of five seeds, and its FHLR is higher by $\PdNsXgbOFhlr$ points ([$\PdNsXgbOFhlrLo$, $\PdNsXgbOFhlrHi$]), with \nesy{} lower in only \PdNsXgbOFhlrWins\ seed.
It is more accurate than logistic regression with $\mathbf{s}$ ($\PdNsLrOAcc$ points, all five seeds) but not safer (FHLR difference $\PdNsLrOFhlr$ points, lower in \PdNsLrOFhlrWins\ seed).
The re-run reproduces logistic regression with $\mathbf{s}$ exactly (\ReLrOAccPm\,\%) and XGBoost with $\mathbf{s}$ within 0.2--0.4 points (accuracy \ReXgbOAccPm\,\% against the original \MsXgbOAccPm\,\%, FHLR \ReXgbOFhlrPm\,\% against \MsXgbOFhlrPm\,\%).
The data therefore support that clause-scoped symbolic features account for most of the improvement over text-only models; they do not support that the neural architecture improves on gradient boosting given the same features.

\subsection{Ablation}
\begin{table*}[!t]
\caption{Ablation of \nesy{} (mean $\pm$ SD over five seeds). $^\dagger$Also trained without $\ahr$, on the plain text with $L=112$. $^\ddagger$Plain text, $L=96$, learning rate $2.5\times10^{-5}$, no symbolic stream. The rows differ in more than one factor, so this is not a clean single-factor study.}
\label{tab:ablation}
\centering\footnotesize
\input{tables/tab_ablation}
\end{table*}
Table~\ref{tab:ablation} gives the ablations.
Removing the asymmetric penalty \emph{improves} the mean accuracy from \MsNsAccPm\,\% to \AbNoAhrAccPm\,\%, and in none of the five seeds is the full model better ($\PdNsNoAhrAcc$ points, [$\PdNsNoAhrAccLo$, $\PdNsNoAhrAccHi$]); the FHLR is essentially unchanged (\AbNoAhrFhlrPm\,\% against \MsNsFhlrPm\,\%).
With the fixed, untuned weights ($\alpha_1=0.65$, $\alpha_2=0.40$, $\delta=0.5$), the hinge term did not reduce test leakage; one plausible reason is that it also penalises the roughly 3.6\,\% noisy training items, which we did not test.
Removing cross-attention (together with the clause-marked input and the loss, see the table note) lowers the accuracy to \AbNoCaAccPm\,\%, \PdNoAhrNoCaAcc\ points below the variant without $\ahr$ ([$\PdNoAhrNoCaAccLo$, $\PdNoAhrNoCaAccHi$], all five seeds), and raises the FHLR to \AbNoCaFhlrPm\,\%; because several factors change at once the effect cannot be attributed to the attention alone.
Removing the symbolic stream altogether gives Vanilla DistilBERT (\MsDbAccPm\,\%).

\subsection{High-risk additive subsets}
\begin{table*}[!t]
\caption{Accuracy / false-Halal leakage (\%) on test items that mention a sensitive additive (seed 42). $N$ is the subset size. LR: logistic regression, RF: random forest, XGB: XGBoost, all on TF-IDF features.}
\label{tab:ecode}
\centering\footnotesize
\input{tables/tab_ecode}
\end{table*}
Table~\ref{tab:ecode} breaks the seed-42 results down by additive: carmine (E120, $N=\EcCarmN$), mono- and diglycerides (E471, $N=\EcGlycN$), glycerol (E422, $N=\EcGlyolN$) and gelatine (E441, $N=\EcGelN$).
On the largest subset, E471, text-only models leak \EcGlycTxtFhlrMin--\EcGlycTxtFhlrMax\,\% of the risky items into Halal, \nesy{} \EcGlycNsFhlr\,\% at \EcGlycNsAcc\,\% accuracy.
On E422, where synonyms (glycerine, propanetriol) and cross-clause qualifiers co-occur, leakage is \EcGlyolTxtFhlrMin--\EcGlyolTxtFhlrMax\,\% for text-only models, \EcGlyolDbFhlr\,\% for Vanilla DistilBERT and \EcGlyolNsFhlr\,\% for \nesy{}.
The subsets are small and come from one seed: one item changes the accuracy by $100/N$, i.e.\ by 5.0 points on E120, and on the carmine subset text-only XGBoost makes no error (\EcCarmXgbAcc\,\%) whereas \nesy{} makes one (\EcCarmNsAcc\,\%).

\subsection{Counterfactual modifier-swap suite}
\begin{figure}[!t]
\centering
\includegraphics[width=\columnwidth]{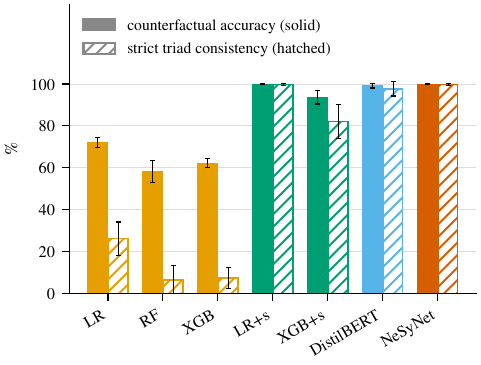}
\caption{Counterfactual accuracy and strict triad consistency (mean $\pm$ SD over five seeds).}
\label{fig:cf}
\end{figure}

\begin{table}[!t]
\caption{Counterfactual suite (\NCf\ items, \NTriplets\ triplets; mean $\pm$ SD over five seeds). Triad consistency requires all three members of a triplet to be predicted correctly; CF FHLR is the false-Halal leakage on the suite. The two ``$+\,\mathbf{s}$'' rows are the values of the original run. LR: logistic regression, RF: random forest, XGB: XGBoost; $\mathbf{s}$: symbolic vector.}
\label{tab:cf}
\centering\scriptsize
\setlength{\tabcolsep}{3.5pt}
\input{tables/tab_cf}
\end{table}
Table~\ref{tab:cf} and Fig.~\ref{fig:cf} report the suite, in which only the clause-local modifier changes within a triplet.
The text-only baselines fail it: counterfactual accuracy of \CfTxtAccMin--\CfTxtAccMax\,\%, strict triad consistency of \CfTxtTriadMin--\CfTxtTriadMax\,\% and counterfactual FHLR of \CfTxtLeakMin--\CfTxtLeakMax\,\%.
This is expected from the construction of the items (Section~\ref{sec:data}), where the n-gram content is nearly unchanged by the swap.
Vanilla DistilBERT, which sees word order, reaches \CfDbAccPm\,\% and \CfDbTriadPm\,\% triad consistency; \nesy{} reaches \CfNsAccPm\,\% and \CfNsTriadPm\,\% with no counterfactual leakage.
Logistic regression with $\mathbf{s}$ obtains the same values as \nesy{} (\CfLrOAccPm\,\% and \CfLrOTriadPm\,\%), and XGBoost with $\mathbf{s}$ reaches \CfXgbOAccPm\,\% and \CfXgbOTriadPm\,\% (\ReXgbOTriad\,\% in the re-run).
Because $\mathbf{s}$ is computed from the unperturbed declaration, the suite largely tests whether the symbolic vector is present.

\subsection{Three-lane receiving triage}
\label{sec:triage}
\begin{table}[!t]
\caption{Three-lane triage at $\tau=0.85$ (mean $\pm$ SD over five seeds). Rows with $\mathbf{s}$ are from the re-run of the ontology-augmented baselines. LR: logistic regression, RF: random forest, XGB: XGBoost; $\mathbf{s}$: symbolic vector.}
\label{tab:triage}
\centering\scriptsize
\setlength{\tabcolsep}{3.5pt}
\input{tables/tab_triage}
\end{table}
Table~\ref{tab:triage} simulates receiving-dock routing.
\nesy{} auto-clears \TrNsGreenPm\,\% of the incoming items into the green lane with a green-lane contamination of \TrNsContPm\,\%; since Halal items are \PctHalal\,\% of the benchmark, this is close to the largest share the green lane can hold.
Text-only logistic regression and random forest reach zero contamination only by auto-clearing \TrLrGreen\,\% and \TrRfGreen\,\% of the items and sending \TrLrManual\,\% and \TrRfManual\,\% to audit; text-only XGBoost auto-clears \TrXgbGreenPm\,\% at \TrXgbContPm\,\% contamination.
Models with $\mathbf{s}$ lie close to \nesy{}: logistic regression with $\mathbf{s}$ clears \TrLrOGreenPm\,\% at \TrLrOContPm\,\%, XGBoost with $\mathbf{s}$ \TrXgbOGreenPm\,\% at \TrXgbOContPm\,\%, and Vanilla DistilBERT \TrDbGreenPm\,\% at \TrDbContPm\,\%.
Differences in contamination among these four are of the size of their standard deviations, so the simulation does not separate them; it shows how a confidence threshold turns a classifier into a receiving rule.

\input{sections/lto}

%% file: tables/tab_main.tex
\begin{tabular}{@{}lccccc@{}}
\toprule
Model & Acc.\ (\%) & Macro-F1 (\%) & ROC-AUC (\%) & FHLR (\%) $\downarrow$ & Haram$\to$Halal (\%) $\downarrow$ \\
\midrule
Logistic Regression (TF-IDF) & 86.62 $\pm$ 1.45 & 87.81 $\pm$ 1.38 & 97.13 $\pm$ 0.63 & 13.42 $\pm$ 2.72 & 3.71 $\pm$ 2.51 \\
Random Forest (TF-IDF) & 86.25 $\pm$ 1.39 & 86.27 $\pm$ 1.45 & 96.22 $\pm$ 0.56 & 21.73 $\pm$ 1.92 & 6.11 $\pm$ 3.22 \\
XGBoost (TF-IDF) & 86.71 $\pm$ 1.45 & 87.64 $\pm$ 1.41 & 96.83 $\pm$ 0.45 & 15.24 $\pm$ 1.54 & \textbf{2.18 $\pm$ 1.54} \\
\midrule
Logistic Regression (TF-IDF + $\mathbf{s}$) & 94.54 $\pm$ 0.85 & 94.29 $\pm$ 0.89 & 99.42 $\pm$ 0.12 & \textbf{3.46 $\pm$ 0.31} & 2.62 $\pm$ 1.25 \\
XGBoost (TF-IDF + $\mathbf{s}$) & 96.25 $\pm$ 1.24 & 96.24 $\pm$ 1.33 & 99.66 $\pm$ 0.15 & 3.81 $\pm$ 0.83 & 2.83 $\pm$ 2.11 \\
\midrule
Vanilla DistilBERT & 93.71 $\pm$ 0.35 & 93.19 $\pm$ 0.21 & 99.21 $\pm$ 0.23 & 6.58 $\pm$ 1.85 & 5.24 $\pm$ 2.08 \\
\textsc{Halal-NeSyNet} (ours) & \textbf{97.04 $\pm$ 0.34} & \textbf{96.90 $\pm$ 0.35} & \textbf{99.84 $\pm$ 0.08} & 4.68 $\pm$ 1.16 & 5.24 $\pm$ 2.36 \\
\bottomrule
\end{tabular}

%% file: tables/tab_perclass.tex
\begin{tabular}{@{}lcccccc@{}}
\toprule
Class & Support & LR & RF & XGB & DistilBERT & \textsc{NeSyNet} \\
\midrule
Halal & 249 & 88.45 & 89.80 & 89.41 & 95.07 & \textbf{97.05} \\
Non-Halal & 92 & 95.45 & 93.18 & \textbf{97.78} & 92.40 & 95.45 \\
Mushbooh & 139 & 80.85 & 80.82 & 81.48 & 92.20 & \textbf{96.73} \\
\bottomrule
\end{tabular}

%% file: tables/tab_paired.tex
\begin{tabular}{@{}llccc@{}}
\toprule
Comparison & Metric & Mean difference & 95\% CI & Better in \\
\midrule
\textsc{NeSyNet} $-$ Vanilla DistilBERT & Accuracy (pp) & +3.34 & [+2.93, +3.75] & 5/5 seeds \\
\textsc{NeSyNet} $-$ Vanilla DistilBERT & FHLR (pp) & -1.90 & [-4.58, +0.77] & 4/5 seeds \\
\textsc{NeSyNet} $-$ XGBoost (TF-IDF) & Accuracy (pp) & +10.33 & [+8.13, +12.54] & 5/5 seeds \\
\textsc{NeSyNet} $-$ XGBoost (TF-IDF) & FHLR (pp) & -10.56 & [-13.02, -8.11] & 5/5 seeds \\
\textsc{NeSyNet} $-$ XGBoost (TF-IDF + $\mathbf{s}$) & Accuracy (pp) & +0.59 & [-1.11, +2.28] & 3/5 seeds \\
\textsc{NeSyNet} $-$ XGBoost (TF-IDF + $\mathbf{s}$) & FHLR (pp) & +1.21 & [-0.28, +2.71] & 1/5 seeds \\
\textsc{NeSyNet} $-$ LR (TF-IDF + $\mathbf{s}$) & Accuracy (pp) & +2.50 & [+1.54, +3.47] & 5/5 seeds \\
\textsc{NeSyNet} $-$ LR (TF-IDF + $\mathbf{s}$) & FHLR (pp) & +1.21 & [-0.23, +2.65] & 1/5 seeds \\
full $-$ w/o $\mathcal{L}_{\mathrm{AHR}}$ & Accuracy (pp) & -0.50 & [-1.06, +0.07] & 0/5 seeds \\
full $-$ w/o $\mathcal{L}_{\mathrm{AHR}}$ & FHLR (pp) & +0.09 & [-0.36, +0.54] & 1/5 seeds \\
w/o $\mathcal{L}_{\mathrm{AHR}}$ $-$ w/o cross-attention$^{\dagger}$ & Accuracy (pp) & +1.29 & [+0.39, +2.19] & 5/5 seeds \\
w/o $\mathcal{L}_{\mathrm{AHR}}$ $-$ w/o cross-attention$^{\dagger}$ & FHLR (pp) & -1.82 & [-3.45, -0.19] & 4/5 seeds \\
\bottomrule
\end{tabular}

%% file: tables/tab_ablation.tex
\begin{tabular}{@{}lcccc@{}}
\toprule
Variant & Acc.\ (\%) & Macro-F1 (\%) & FHLR (\%) $\downarrow$ & Haram$\to$Halal (\%) $\downarrow$ \\
\midrule
\textsc{Halal-NeSyNet} (full) & 97.04 $\pm$ 0.34 & 96.90 $\pm$ 0.35 & 4.68 $\pm$ 1.16 & 5.24 $\pm$ 2.36 \\
w/o $\mathcal{L}_{\mathrm{AHR}}$ (weighted CE) & \textbf{97.54 $\pm$ 0.40} & \textbf{97.47 $\pm$ 0.44} & \textbf{4.59 $\pm$ 1.09} & \textbf{4.58 $\pm$ 2.08} \\
w/o cross-attention$^{\dagger}$ & 96.25 $\pm$ 0.87 & 96.03 $\pm$ 0.99 & 6.41 $\pm$ 1.24 & 5.89 $\pm$ 2.12 \\
w/o symbolic stream (= Vanilla DistilBERT)$^{\ddagger}$ & 93.71 $\pm$ 0.35 & 93.19 $\pm$ 0.21 & 6.58 $\pm$ 1.85 & 5.24 $\pm$ 2.08 \\
\bottomrule
\end{tabular}

%% file: tables/tab_ecode.tex
\begin{tabular}{@{}lcccccc@{}}
\toprule
Additive & $N$ & LR & RF & XGB & DistilBERT & \textsc{NeSyNet} \\
\midrule
E120 & 20 & 95.0 / 5.0 & 85.0 / 15.0 & 100.0 / 0.0 & 85.0 / 10.0 & 95.0 / 0.0 \\
E471 & 143 & 90.2 / 8.3 & 91.6 / 10.7 & 93.0 / 7.1 & 97.9 / 2.4 & 98.6 / 1.2 \\
E422 & 38 & 79.0 / 18.8 & 89.5 / 25.0 & 84.2 / 12.5 & 89.5 / 18.8 & 92.1 / 6.2 \\
E441 & 39 & 89.7 / 4.3 & 84.6 / 17.4 & 92.3 / 4.3 & 92.3 / 8.7 & 97.4 / 4.3 \\
\bottomrule
\end{tabular}

%% file: tables/tab_cf.tex
\begin{tabular}{@{}lccc@{}}
\toprule
Model & \shortstack{CF acc.\\(\%)} & \shortstack{Triad cons.\\(\%)} & \shortstack{CF FHLR\\(\%) $\downarrow$} \\
\midrule
LR (TF-IDF) & 71.89 $\pm$ 2.38 & 26.00 $\pm$ 8.05 & 26.83 $\pm$ 4.10 \\
RF (TF-IDF) & 58.07 $\pm$ 5.28 & 6.33 $\pm$ 6.91 & 28.67 $\pm$ 8.55 \\
XGB (TF-IDF) & 62.15 $\pm$ 2.17 & 7.33 $\pm$ 4.94 & 21.17 $\pm$ 3.26 \\
\midrule
LR (TF-IDF + $\mathbf{s}$) & \textbf{99.89 $\pm$ 0.17} & \textbf{99.67 $\pm$ 0.50} & \textbf{0.00 $\pm$ 0.00} \\
XGB (TF-IDF + $\mathbf{s}$) & 93.59 $\pm$ 3.16 & 82.11 $\pm$ 8.05 & \textbf{0.00 $\pm$ 0.00} \\
\midrule
Vanilla DistilBERT & 99.22 $\pm$ 1.15 & 97.67 $\pm$ 3.45 & 1.17 $\pm$ 1.73 \\
\textsc{Halal-NeSyNet} & \textbf{99.89 $\pm$ 0.17} & \textbf{99.67 $\pm$ 0.50} & \textbf{0.00 $\pm$ 0.00} \\
\bottomrule
\end{tabular}

%% file: tables/tab_triage.tex
\begin{tabular}{@{}lcc@{}}
\toprule
Model & \shortstack{Auto-cleared to\\green lane (\%)} & \shortstack{Green-lane\\contamination (\%) $\downarrow$} \\
\midrule
LR (TF-IDF) & 22.25$\pm$1.85 & 0.00$\pm$0.00 \\
RF (TF-IDF) & 18.63$\pm$1.26 & 0.00$\pm$0.00 \\
XGB (TF-IDF) & 35.96$\pm$2.59 & 4.11$\pm$1.68 \\
\midrule
LR (TF-IDF + $\mathbf{s}$) & 42.96$\pm$0.86 & 0.96$\pm$0.75 \\
XGB (TF-IDF + $\mathbf{s}$) & 48.54$\pm$0.91 & 1.46$\pm$0.48 \\
\midrule
Vanilla DistilBERT & 47.13$\pm$2.63 & 1.26$\pm$1.43 \\
\textsc{Halal-NeSyNet} & 49.33$\pm$0.71 & 0.76$\pm$0.53 \\
\bottomrule
\end{tabular}

%% file: sections/lto.tex
\subsection{Leave-one-template-family-out}
\label{sec:lto}
The random split of Section~\ref{sec:data} leaves every template family in the training data.
To test generalisation to an unseen template, we hold out one of the ten generator families (eight additive templates and two negation templates) in turn as the test set (\LtoTestMin--\LtoTestMax\ items) and train on the real records and the other nine families, with an 85/15 split of that pool into about \LtoTrainMean\ training and \LtoValMean\ validation items.
All models, hyper-parameters and the selection of the best epoch are unchanged (\nesy{}, \nesy{} without $\ahr$, Vanilla DistilBERT and the five classical models), with \LtoNSeeds\ seeds per fold.
The unit of analysis is the family: seeds are averaged within a family, and means, standard deviations and paired differences use the \LtoNFolds\ families ($t$ interval with 9 degrees of freedom).
The counterfactual triplets of the held-out family, available for six of the ten families, are evaluated as an unseen-family scope test.
The symbolic vector is still computed from the unperturbed declaration (Section~\ref{sec:perturb}), so this protocol tests template generalisation with an oracle-quality parser, not robustness to parsing errors.

\begin{figure*}[!t]
\centering
\includegraphics[width=0.94\textwidth]{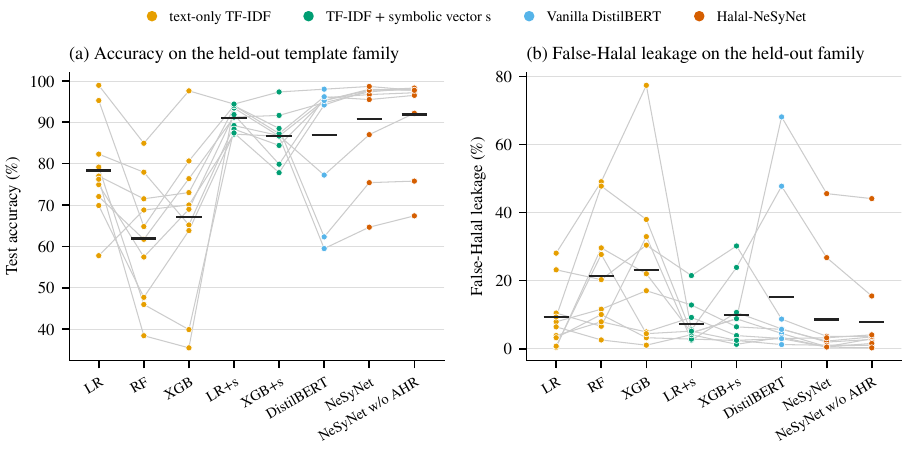}
\caption{Leave-one-template-family-out. (a) Test accuracy and (b) false-Halal leakage on the held-out family for each model; each dot is one family (mean of the seeds), grey lines join the same family and the black bar is the mean over the ten families.}
\label{fig:lto}
\end{figure*}

\begin{table*}[!t]
\caption{Leave-one-template-family-out results (mean $\pm$ SD over the ten held-out families, seeds averaged within a family). CF triad: strict triad consistency on the counterfactual triplets of the held-out family (six families).}
\label{tab:lto}
\centering\footnotesize
\input{tables/tab_lto}
\end{table*}
Table~\ref{tab:lto} and Fig.~\ref{fig:lto} give the results; Table~\ref{tab:ltofold} lists every family.
Holding out a family lowers every model relative to the random split.
The text-only models fall to \LtoLrAccPm\,\% (logistic regression), \LtoRfAccPm\,\% (random forest) and \LtoXgbAccPm\,\% (XGBoost), from \TxtAccMin--\TxtAccMax\,\% on the random split, and their counterfactual triad consistency is at most \LtoTxtCfTriadMax\,\%.
The models with the symbolic vector $\mathbf{s}$ degrade much less: logistic regression \LtoLrOAccPm\,\% (between \LtoLrOAccMin\,\% and \LtoLrOAccMax\,\% on every family) and XGBoost \LtoXgbOAccPm\,\%.
Vanilla DistilBERT reaches \LtoDbAccPm\,\%, \nesy{} \LtoNsAccPm\,\% and \nesy{} without $\ahr$ \LtoNoAhrAccPm\,\%.

\begin{table*}[!t]
\caption{Test accuracy (\%) on each held-out template family (mean of \LtoNSeeds\ seeds); best per family in bold.}
\label{tab:ltofold}
\centering\footnotesize
\input{tables/tab_lto_fold}
\end{table*}

\begin{table*}[!t]
\caption{Family-paired differences over the ten held-out families (first model minus second; mean, $t$-based 95\,\% interval, number of families in which the first model is better).}
\label{tab:ltopaired}
\centering\footnotesize
\input{tables/tab_lto_paired}
\end{table*}
Table~\ref{tab:ltopaired} compares the models family by family.
\nesy{} is more accurate than Vanilla DistilBERT by $\LtoPdNsDbAcc$ points ([$\LtoPdNsDbAccLo$, $\LtoPdNsDbAccHi$]) and its false-Halal leakage differs by $\LtoPdNsDbFhlr$ points ([$\LtoPdNsDbFhlrLo$, $\LtoPdNsDbFhlrHi$]), both in 9 of 10 families, so the symbolic vector helps the neural model to generalise.
Against the classical models with $\mathbf{s}$ the difference cannot be resolved with ten families: $\LtoPdNsXgbOAcc$ points ([$\LtoPdNsXgbOAccLo$, $\LtoPdNsXgbOAccHi$]) against XGBoost and $\LtoPdNsLrOAcc$ points ([$\LtoPdNsLrOAccLo$, $\LtoPdNsLrOAccHi$]) against logistic regression.
The spread, however, is very different: the accuracy of \nesy{} ranges from \LtoNsAccMin\,\% to \LtoNsAccMax\,\% across families (standard deviation \LtoNsAccSd\ points) whereas that of logistic regression with $\mathbf{s}$ stays within \LtoLrOAccMin--\LtoLrOAccMax\,\% (\LtoLrOAccSd\ points).
Removing $\ahr$ again does not hurt: the full model is $\LtoPdNsNoAhrAcc$ points relative to the variant without it ([$\LtoPdNsNoAhrAccLo$, $\LtoPdNsNoAhrAccHi$]), better in only \LtoPdNsNoAhrAccWins\ of 10 families.

The neural models fail on two families, in opposite directions.
When the gelatine template (E441) is held out, \nesy{} predicts Non-Halal for \LtoGelNsHalalToHaram\,\% of the truly Halal items (Vanilla DistilBERT \LtoGelDbHalalToHaram\,\%, logistic regression with $\mathbf{s}$ \LtoGelLrOHalalToHaram\,\%), an over-cautious error.
When the ginger-beer negation template is held out, \nesy{} predicts Halal for \LtoGingNsMushToHalal\,\% of the Mushbooh items and Vanilla DistilBERT for \LtoGingDbMushToHalal\,\%, against \LtoGingLrOMushToHalal\,\% for logistic regression with $\mathbf{s}$, the unsafe direction; a plausible reason, which we did not test, is that words such as ``non-alcoholic'' and ``dealcoholized'' in the additive clause of these templates push a text encoder towards Halal.
Pooled over all held-out items, \nesy{} lets \LtoAllNsMushToHalal\,\% of the Mushbooh items pass as Halal, Vanilla DistilBERT \LtoAllDbMushToHalal\,\%, logistic regression with $\mathbf{s}$ \LtoAllLrOMushToHalal\,\% and XGBoost with $\mathbf{s}$ \LtoAllXgbOMushToHalal\,\%.
On the counterfactual triplets of the held-out family (six families), strict triad consistency is \LtoNsCfTriadPm\,\% for \nesy{}, \LtoLrOCfTriadPm\,\% for logistic regression with $\mathbf{s}$, \LtoXgbOCfTriadPm\,\% for XGBoost with $\mathbf{s}$ and \LtoDbCfTriadPm\,\% for Vanilla DistilBERT.

%% file: tables/tab_lto.tex
\begin{tabular}{@{}lccccc@{}}
\toprule
Model & Acc.\ (\%) & Macro-F1 (\%) & FHLR (\%) $\downarrow$ & Haram$\to$Halal (\%) $\downarrow$ & CF triad (\%) \\
\midrule
Logistic Regression (TF-IDF) & 78.35 $\pm$ 11.92 & 79.94 $\pm$ 11.30 & 9.35 $\pm$ 9.29 & 10.86 $\pm$ 15.38 & 2.96 $\pm$ 5.38 \\
Random Forest (TF-IDF) & 61.91 $\pm$ 14.77 & 60.23 $\pm$ 14.82 & 21.32 $\pm$ 16.84 & 19.02 $\pm$ 17.34 & 0.00 $\pm$ 0.00 \\
XGBoost (TF-IDF) & 67.10 $\pm$ 18.26 & 63.54 $\pm$ 18.18 & 23.14 $\pm$ 23.39 & 30.68 $\pm$ 32.67 & 0.00 $\pm$ 0.00 \\
\midrule
Logistic Regression (TF-IDF + $\mathbf{s}$) & 91.11 $\pm$ 2.88 & 90.75 $\pm$ 3.17 & \textbf{7.30 $\pm$ 5.88} & 6.80 $\pm$ 9.63 & 84.44 $\pm$ 28.80 \\
XGBoost (TF-IDF + $\mathbf{s}$) & 86.70 $\pm$ 5.50 & 84.66 $\pm$ 8.15 & 9.87 $\pm$ 9.69 & 12.38 $\pm$ 18.64 & 52.96 $\pm$ 47.65 \\
\midrule
Vanilla DistilBERT & 86.85 $\pm$ 14.89 & 85.07 $\pm$ 17.09 & 15.10 $\pm$ 23.18 & 7.76 $\pm$ 12.88 & 79.63 $\pm$ 38.17 \\
\textsc{Halal-NeSyNet} (ours) & 90.87 $\pm$ 11.76 & 90.10 $\pm$ 12.71 & 8.59 $\pm$ 15.24 & \textbf{3.57 $\pm$ 6.79} & 92.41 $\pm$ 13.08 \\
\textsc{Halal-NeSyNet} w/o $\mathcal{L}_{\mathrm{AHR}}$ & \textbf{91.88 $\pm$ 11.03} & \textbf{91.07 $\pm$ 12.07} & 7.92 $\pm$ 13.42 & 4.96 $\pm$ 8.88 & 93.52 $\pm$ 13.34 \\
\bottomrule
\end{tabular}

%% file: tables/tab_lto_fold.tex
\begin{tabular}{@{}lcccccccc@{}}
\toprule
Held-out family & LR & RF & XGB & LR+s & XGB+s & DistilBERT & NeSyNet & NeSyNet w/o AHR \\
\midrule
E471 & 72.0 & 61.7 & 76.4 & 91.2 & 91.7 & 94.7 & 97.4 & \textbf{98.2} \\
E472e & 82.3 & 77.9 & 65.2 & 89.2 & 86.8 & 77.2 & 87.0 & \textbf{92.2} \\
E441 & 57.8 & 68.8 & 70.0 & \textbf{87.1} & 86.6 & 59.5 & 64.6 & 67.4 \\
E422 & 79.2 & 46.0 & 39.9 & 88.3 & 84.4 & 95.2 & \textbf{98.0} & 97.7 \\
rennet & 69.9 & 47.7 & 63.8 & 87.4 & 77.8 & 94.2 & 97.7 & \textbf{97.9} \\
E920 & 74.9 & 57.4 & 69.0 & 91.9 & 79.9 & 95.9 & 96.7 & \textbf{97.1} \\
E433 & 77.0 & 71.5 & 73.0 & 93.4 & 86.7 & 95.3 & 97.7 & \textbf{98.3} \\
E631 & 76.2 & 38.4 & 35.5 & 94.0 & 87.3 & 96.2 & 95.5 & \textbf{96.5} \\
ginger beer & \textbf{95.3} & 64.8 & 80.6 & 94.0 & 88.5 & 62.3 & 75.4 & 75.8 \\
wine vinegar & \textbf{98.9} & 84.9 & 97.6 & 94.4 & 97.3 & 98.0 & 98.7 & 97.7 \\
\bottomrule
\end{tabular}

%% file: tables/tab_lto_paired.tex
\begin{tabular}{@{}llccc@{}}
\toprule
Comparison & Metric & Mean difference & 95\% CI & Better in \\
\midrule
\textsc{NeSyNet} $-$ Vanilla DistilBERT & Accuracy (pp) & +4.02 & [+0.94, +7.10] & 9/10 families \\
\textsc{NeSyNet} $-$ Vanilla DistilBERT & FHLR (pp) & -6.51 & [-12.39, -0.63] & 9/10 families \\
\textsc{NeSyNet} $-$ XGBoost (TF-IDF + $\mathbf{s}$) & Accuracy (pp) & +4.18 & [-5.27, +13.63] & 8/10 families \\
\textsc{NeSyNet} $-$ XGBoost (TF-IDF + $\mathbf{s}$) & FHLR (pp) & -1.28 & [-12.51, +9.95] & 8/10 families \\
\textsc{NeSyNet} $-$ LR (TF-IDF + $\mathbf{s}$) & Accuracy (pp) & -0.23 & [-8.34, +7.87] & 7/10 families \\
\textsc{NeSyNet} $-$ LR (TF-IDF + $\mathbf{s}$) & FHLR (pp) & +1.30 & [-10.94, +13.54] & 8/10 families \\
full $-$ w/o $\mathcal{L}_{\mathrm{AHR}}$ & Accuracy (pp) & -1.01 & [-2.25, +0.24] & 2/10 families \\
full $-$ w/o $\mathcal{L}_{\mathrm{AHR}}$ & FHLR (pp) & +0.68 & [-2.09, +3.44] & 5/10 families \\
\textsc{NeSyNet} $-$ XGBoost (TF-IDF) & Accuracy (pp) & +23.78 & [+7.21, +40.35] & 8/10 families \\
\textsc{NeSyNet} $-$ XGBoost (TF-IDF) & FHLR (pp) & -14.55 & [-27.37, -1.73] & 9/10 families \\
XGBoost (TF-IDF + $\mathbf{s}$) $-$ DistilBERT & Accuracy (pp) & -0.16 & [-11.65, +11.33] & 3/10 families \\
XGBoost (TF-IDF + $\mathbf{s}$) $-$ DistilBERT & FHLR (pp) & -5.23 & [-20.91, +10.45] & 4/10 families \\
\bottomrule
\end{tabular}

%% file: sections/discussion.tex
\section{Discussion and Limitations}
\label{sec:discussion}

\subsection{What the experiments show}
Four results stand out.
First, clause scope matters, and it is the symbolic features that deliver it: text-only bag-of-words models are at \TxtAccMin--\TxtAccMax\,\% accuracy and at \CfTxtTriadMin--\CfTxtTriadMax\,\% triad consistency, a failure that the construction of the items makes likely; Vanilla DistilBERT, which sees word order, recovers much of it (\CfDbTriadPm\,\% triad consistency); and every model that receives the clause-parsed vector $\mathbf{s}$ reaches \MsLrOAcc\,\% accuracy or more.
Second, the neural architecture is not what separates the best models: XGBoost with $\mathbf{s}$ cannot be distinguished from \nesy{} in accuracy at this sample size (paired difference $\PdNsXgbOAcc$ points, interval [$\PdNsXgbOAccLo$, $\PdNsXgbOAccHi$]) and has lower leakage in most seeds, at a small fraction of the cost.
The mean accuracy of \nesy{} is the highest and its seed-to-seed spread the smallest among the models that receive $\mathbf{s}$ (\MsNsAccSd\ against \MsXgbOAccSd\ points for XGBoost with $\mathbf{s}$), which may matter operationally, but we did not test this further.
Third, the asymmetric risk loss, designed for the safety metric, did not reduce leakage and lowered accuracy in all five seeds, and again in eight of ten held-out families (Section~\ref{sec:lto}).
Fourth, the leave-one-template-family-out evaluation preserves the ranking but exposes a difference in robustness: the models with $\mathbf{s}$ keep \LtoLrOAccMin--\LtoLrOAccMax\,\% (logistic regression) on every unseen family, whereas \nesy{} ranges from \LtoNsAccMin\,\% to \LtoNsAccMax\,\%, with its two failures in opposite directions (over-cautious on the gelatine family, unsafe on the ginger-beer family); the network adds variance across templates, not accuracy over a linear model with the same features.
Safety, measured by false-Halal leakage, is not improved by the largest model: the lowest FHLR belongs to logistic regression with $\mathbf{s}$ (\MsLrOFhlrPm\,\%), and \nesy{} misses \CmNsHaramToHalal\ of \SupHaram\ true Non-Halal items in the seed-42 split.

\subsection{Why the benchmark favours symbolic features}
The labels and $\mathbf{s}$ are produced by the same rule engine, and the text perturbations (Section~\ref{sec:perturb}) are applied afterwards, so $\mathbf{s}$ is an oracle-like reader of the unperturbed declaration while the text-only models read a degraded version.
High accuracy of the models with $\mathbf{s}$ is therefore evidence that they can use the rule outputs, not that they have learned Halal jurisprudence; the benchmark tests fidelity to a documented rule policy.
This is the familiar situation in which a model can score well on a data set by exploiting regularities that do not transfer \cite{geirhos2020shortcut,mccoy2019right,gururangan2018annotation}, and, like heuristic labelling functions \cite{ratner2017snorkel}, the labels are only as good as the rules.
A deployed parser would run on the same noisy text, possibly after OCR, as the encoder, and we did not evaluate that setting.
The counterfactual suite shares templates with the training data and is solved by $\mathbf{s}$ alone, so it should be read as a unit test of scope sensitivity rather than as evidence of generalisation \cite{kaushik2020learning,gardner2020evaluating,ribeiro2020beyond}.

\subsection{Operational reading for the three pillars}
\emph{Certification.} A classifier does not certify; authority remains with accredited bodies.
Because the rule base is decoupled from the network weights through $\mathbf{s}$, auditors can change jurisdiction-specific rules (for example on shellac E904 or carmine E120) without retraining the encoder, although any rule change also changes the labels that the system is measured against.
\emph{Traceability.} The parser can record the clause and the additive that triggered a Mushbooh or Non-Halal flag, which gives a clause-level audit trail to attach to a warehouse or ledger record \cite{rejeb2018halal,tan2022blockchain}.
\emph{Quality assurance.} With a false-Halal leakage of \MsLrOFhlr--\MsNsFhlr\,\% for the best models, unattended intake is not justified; a reject-option rule \cite{chow1970optimum,geifman2017selective} such as the three-lane routing sends doubtful items to a certificate-of-analysis audit, and in our simulation the models with $\mathbf{s}$ and \nesy{} keep green-lane contamination at \TrNsCont--\TrXgbOCont\,\% while auto-clearing \TrLrOGreen--\TrNsGreen\,\% of the cargo.
The threshold was fixed at 0.85 on an uncalibrated softmax confidence and its operating characteristic was not studied.

\subsection{Limitations}
\begin{itemize}
\item \emph{Predominantly synthetic data.} \PctSyn\,\% of the items are generated from ten templates, \NDupText\ texts are repeated, and results on unconstrained real packaging may be lower; per-item predictions were not stored, so we cannot report separate results for the \NReal\ real records.
\item \emph{Rule-derived labels and feature-label coupling.} Labels come from a rule base that no certification body has audited (Section~\ref{sec:labels}); the symbolic vector is computed by the same rules and is not degraded by the text perturbations.
\item \emph{Counterfactual suite and template generalisation.} The suite is template-sharing and largely solvable from $\mathbf{s}$ alone; the leave-one-template-family-out evaluation holds out whole families, but the ten families come from one generator, use three seeds, and the symbolic vector is still computed from the unperturbed text.
\item \emph{Multi-factor ablations.} The ``w/o cross-attention'' variant also changes the loss, the input text and the sequence length, and Vanilla DistilBERT differs in input, length and learning rate.
\item \emph{Untuned loss.} The weights of $\ahr$ and the threshold $\tau$ were fixed a priori; a tuned loss might behave differently, which we did not test.
\item \emph{Few seeds and small subsets.} Five seeds give wide intervals; the per-class table, confusion matrices and additive subsets use one seed, and the subsets have only \EcCarmN\ to \EcGlycN\ items.
\item \emph{Re-run baselines.} Per-seed values of the ontology-augmented baselines were reconstructed with a different library version; logistic regression reproduces exactly, XGBoost within about 0.2 points.
\item \emph{Latency} compares a CPU with a GPU and is not a like-for-like measure.
\item \emph{Scope.} English declarations only, no OCR, no multilingual input and no human baseline; the manuscript has not been peer reviewed.
\end{itemize}
Future work should build a benchmark of real products labelled independently by certification auditors, evaluate the parser on OCR output so that the perturbations reach the features, hold out families from an independent generator, tune and calibrate the loss and the routing threshold, and extend the rule base to multilingual declarations.

%% file: sections/conclusion.tex
\section{Conclusion}
\label{sec:conclusion}

We studied three-way (Halal, Non-Halal, Mushbooh) auditing of food ingredient declarations under cross-clause modifier confounding, where a qualifier on one clause must not clear an additive in another.
On a \NItemsFmt-item rule-labelled benchmark and a \NTriplets-triplet counterfactual suite, five-seed results show that text-only TF-IDF classifiers reach \TxtAccMin--\TxtAccMax\,\% accuracy with \TxtFhlrMin--\TxtFhlrMax\,\% false-Halal leakage, that fine-tuned DistilBERT reaches \MsDbAccPm\,\%, and that \nesy{}, which couples DistilBERT to a rule-derived symbolic vector through cross-attention, reaches \MsNsAccPm\,\% accuracy and \CfNsTriadPm\,\% counterfactual triad consistency.
Controlled comparisons temper this: gradient boosting given the same symbolic vector cannot be distinguished from \nesy{} in accuracy and leaks less in most seeds, and the asymmetric risk loss lowered accuracy without reducing leakage; the clause-scoped symbolic features, not the network, account for most of the gain.
Holding out whole template families preserves the ranking (text-only models \LtoTxtAccMin--\LtoTxtAccMax\,\%, \nesy{} \LtoNsAccPm\,\%, logistic regression with the symbolic vector \LtoLrOAccPm\,\%) but shows that \nesy{} is far less stable across templates than the linear model with the same features.
A three-lane receiving simulation shows how a confidence threshold converts any of the strong models into a rule that auto-clears roughly half of the cargo at about one per cent green-lane contamination.
Because labels and features come from one rule engine and most items are templated, the results measure fidelity to a documented policy, not Halal certification; independently labelled real products and parsing of OCR output are the next steps.